\def\year{2021}\relax
\documentclass[letterpaper]{article} % DO NOT CHANGE THIS
\usepackage{aaai20}  % DO NOT CHANGE THIS
\usepackage[utf8]{inputenc}
\usepackage{times}  % DO NOT CHANGE THIS
\usepackage{helvet} % DO NOT CHANGE THIS
\usepackage{courier}  % DO NOT CHANGE THIS
\usepackage[hyphens]{url}  % DO NOT CHANGE THIS
\usepackage{graphicx} % DO NOT CHANGE THIS
\usepackage{graphicx}  % DO NOT CHANGE THIS
\usepackage{amsmath}
\usepackage{booktabs}
\usepackage{xcolor}
\usepackage{tikz}
\usetikzlibrary{arrows}
\usetikzlibrary{arrows.meta}
\usetikzlibrary{decorations.markings}
\usepackage{subcaption}
\usepackage{scalerel}
\usepackage[switch]{lineno}

\usepackage{amsthm}
\usepackage{amsmath}
\usepackage{amssymb}
\usepackage{xspace}
\usepackage{xfrac}

\usetikzlibrary{calc}
\usepackage{relsize}

\tikzset{fontscale/.style = {font=\relsize{#1}}
    }
    
\usepackage{listings}
\usepackage{parcolumns}
\lstdefinestyle{datalogstyle}{
	basicstyle={\codefont\small},
	xleftmargin={14pt},
	numbers=left,
	frame=l,
	stepnumber=1,
	firstnumber=1,
	numberfirstline=true,
	tabsize=2,
	showtabs=false,
	showspaces=false,
	showstringspaces=false,
	extendedchars=true,
	breaklines=true,
	columns=fullflexible,
	keepspaces=true,
	escapeinside={@}{@},
	firstnumber=last,
	captionpos=b,
	commentstyle=\color{black!65},
	numberstyle=\tiny\color{black!65},
	stringstyle=\color{codepurple},
	breakatwhitespace=false, 
	keepspaces=true,                 
	numbersep=5pt,                  
	showspaces=false,                
	showstringspaces=false,
	showtabs=false,
	aboveskip={0.2\baselineskip},
	belowskip={-0.2\baselineskip},
}
\newcommand{\codefont}{\fontfamily{lmtt}\selectfont}
\newcommand{\data}[1]{\texttt{\codefont#1}}
\newcommand{\code}[1]{\data{#1}}

\newcommand{\dep}{\textcolor{brown}{\data{:\!-}}\xspace}
\newcommand{\bluecode}[1]{\textcolor{blue}{\data{#1}}}

\newcommand{\knorf}[1]{\data{Knorf}#1}

\newcommand{\brick}[1]{\scalerel*{\includegraphics{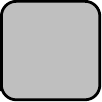}}{B}#1}
\newcommand{\hor}[1]{\scalerel*{\includegraphics{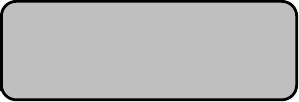}}{B}#1}
\newcommand{\horTwo}[1]{\scalerel*{\includegraphics{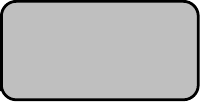}}{B}#1}
\newcommand{\pillarTop}[1]{\scalerel*{\includegraphics{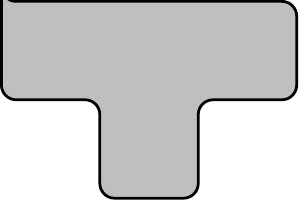}}{B}#1}
\newcommand{\pillarBottom}[1]{\scalerel*{\includegraphics{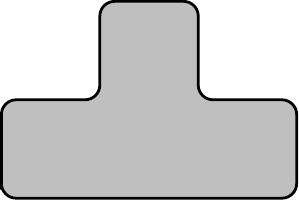}}{B}#1}

\theoremstyle{definition}
\newtheorem{definition}{Definition}

\title{Knowledge Refactoring for Inductive Program Synthesis}
\author{Sebastijan Duman\v{c}i\'{c}\\ % All authors must be in the same font size and format. Use \Large and \textbf to achieve this result when breaking a line
KU Leuven, Belgium \\
\And
Tias Guns \\
KU Leuven, Belgium \\
 \And
 Andrew Cropper \\
 Oxford, United Kingdom
}
 
\begin{document}
 
 %\linenumbers

\maketitle

\begin{abstract}
Humans constantly restructure knowledge to use it more efficiently.
Our goal is to give a machine learning system similar abilities so that it can learn more efficiently.
We introduce the \textit{knowledge refactoring} problem, where the goal is to restructure a learner's knowledge base to reduce its size and to minimise redundancy in it.
We focus on inductive logic programming, where the knowledge base is a logic program.
%  and more specifically on the challenging task of program induction.
% \tias{Smth about why program induction is challenging or hot?}
We introduce \knorf{}, a system which solves the refactoring problem using constraint optimisation.
A key feature of \knorf{} is that, rather than simply removing knowledge, it also introduces new knowledge through \emph{predicate invention}.
% We claim that refactoring a knowledge base can make it more efficient to \emph{learn} from.
We evaluate our approach on two domains: building Lego structures and real-world string transformations.
Our experiments show that learning from refactored knowledge can improve predictive accuracies fourfold and reduce learning times by half.
% \ac{We should probably mention inductive logic programming somewhere.}
\end{abstract}

\section{Introduction}

%Three more points, as a reviewer the negative comments I would have on the paper are:
%
%1. It needs a much clearer problem definition
%2. A discussion about whether the approach is lossless (perhaps a statement about the guarantees)
%3. A clearer definition about what the inputs/parameters of knorf are, and how they influence the solving time.

%People are not mere knowledge accumulators \ac{I think we can cut this sentence. It is implied the following sentences.}.
According to the seminal work of Rumelhart and Norman (\citeyear{Rumelhart1976AccretionTA}), humans exhibit three modes of learning.
\textit{Learning by accretion} is an everyday kind of learning which merely increments a person's knowledge base with new facts.
% , the most common mode,
\textit{Learning by tuning} involves changes in the categories people use for interpreting new information.
% , a more complex mode,
For instance, the process of tuning specialises a child's interpretation of the word `doggie' from all animals to dogs only.
\textit{Restructuring} devises new memory structures and organisation of already stored knowledge, which in turn allows for better accessibility of the acquired knowledge.
This restructuring ability is the most significant mode and is what separates well-performing individuals from others~\cite{Karmiloff1992,Stern2005KnowledgeRA}. %\ac{this ability to restructure knowledge?}.

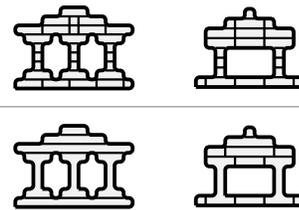
\begin{figure}[t]
	\centering
	\begin{subfigure}{.4\linewidth}
		\begin{flushright}
			\begin{tikzpicture}[scale=0.12]

			\draw[rounded corners=0.5mm,line width=0.5mm, fill=gray, fill opacity=0.1] (0,0)--(0,1)--(1,1)--(1,2)--(2,2)--(2,5)--(1,5)--(1,6)--(0,6)--(0,7)--(3,7)--(3,8)--(5,8)--(5,9)--(8,9)--(8,8)--(10,8)--(10,7)--(13,7)--(13,6)--(12,6)--(12,5)--(11,5)--(11,2)--(12,2)--(12,1)--(13,1)--(13,0)--(0,0);
			
			\draw[rounded corners=0.5mm,line width=0.5mm, fill=white] (3,2)--(4,2)--(4,1)--(5,1)--(5,2)--(6,2)--(6,5)--(5,5)--(5,6)--(4,6)--(4,5)--(3,5)--(3,2);
			
			\draw[rounded corners=0.5mm,line width=0.5mm, fill=white] (7,2)--(8,2)--(8,1)--(9,1)--(9,2)--(10,2)--(10,5)--(9,5)--(9,6)--(8,6)--(8,5)--(7,5)--(7,2);
			
			\draw[line width=0.3mm] (3,0)--(3,1);
			\draw[line width=0.3mm] (6,0)--(6,1);
			\draw[line width=0.3mm] (9,0)--(9,1);
			\draw[line width=0.3mm] (12,0)--(12,1);
			\draw[line width=0.3mm] (1,1)--(12,1);
			
			\draw[line width=0.3mm] (2,2)--(3,2);
			\draw[line width=0.3mm] (6,2)--(7,2);
			\draw[line width=0.3mm] (10,2)--(11,2);
			
			\draw[line width=0.3mm] (2,3)--(3,3);
			\draw[line width=0.3mm] (6,3)--(7,3);
			\draw[line width=0.3mm] (10,3)--(11,3);
			
			\draw[line width=0.3mm] (2,4)--(3,4);
			\draw[line width=0.3mm] (6,4)--(7,4);
			\draw[line width=0.3mm] (10,4)--(11,4);
			
			\draw[line width=0.3mm] (2,5)--(3,5);
			\draw[line width=0.3mm] (6,5)--(7,5);
			\draw[line width=0.3mm] (10,5)--(11,5);
			
			\draw[line width=0.3mm] (1,6)--(12,6);
			\draw[line width=0.3mm] (1,7)--(12,7);
			\draw[line width=0.3mm] (3,8)--(10,8);
			
			\draw[line width=0.3mm] (3,6)--(3,7);
			\draw[line width=0.3mm] (6,6)--(6,7);
			\draw[line width=0.3mm] (7,6)--(7,7);
			\draw[line width=0.3mm] (10,6)--(10,7);
			%\draw[line width=0.3mm] (12,6)--(12,7);
			
			\draw[line width=0.3mm] (6,7)--(6,8);
			\draw[line width=0.3mm] (7,7)--(7,8);

		\end{tikzpicture}
		\end{flushright}
	\end{subfigure}\hspace{2em}
	\begin{subfigure}{.4\linewidth}
		\begin{flushleft}
			\begin{tikzpicture}[scale=0.13]

			\draw[rounded corners=0.5mm,line width=0.5mm, fill=gray, fill opacity=0.1] (0,0)--(11,0)--(11,1)--(9,1)--(9,4)--(10,4)--(10,6)--(8,6)--(8,7)--(6,7)--(6,8)--(5,8)--(5,7)--(3,7)--(3,6)--(1,6)--(1,4)--(2,4)--(2,1)--(0,1)--(0,0);
			
			\draw[rounded corners=0.5mm,line width=0.5mm, fill=white] (3,1)--(8,1)--(8,4)--(7,4)--(4,4)--(3,4)--(3,1);
			
			\draw[line width=0.3mm] (1,0)--(1,1);
			\draw[line width=0.3mm] (4,0)--(4,1);
			\draw[line width=0.3mm] (7,0)--(7,1);
			\draw[line width=0.3mm] (10,0)--(10,1);
			
			\draw[line width=0.3mm] (1,1)--(10,1);
			
			\draw[line width=0.3mm] (2,2)--(3,2);
			\draw[line width=0.3mm] (2,3)--(3,3);
			\draw[line width=0.3mm] (2,4)--(3,4);
			
			\draw[line width=0.3mm] (8,2)--(9,2);
			\draw[line width=0.3mm] (8,3)--(9,3);
			\draw[line width=0.3mm] (8,4)--(9,4);
			
			\draw[line width=0.3mm] (1,5)--(10,5);
			\draw[line width=0.3mm] (1,6)--(10,6);
			\draw[line width=0.3mm] (3,7)--(8,7);
			
			\draw[line width=0.3mm] (4,4)--(4,5);
			\draw[line width=0.3mm] (7,4)--(7,5);
			\draw[line width=0.3mm] (4,5)--(4,6);
			\draw[line width=0.3mm] (7,5)--(7,6);
			\draw[line width=0.3mm] (4,6)--(4,7);
			\draw[line width=0.3mm] (7,6)--(7,7);
		\end{tikzpicture}
		\end{flushleft}

	\end{subfigure}
	
	\vspace{.5em}
	
	\begin{subfigure}{\linewidth}
		\centering
		\begin{tikzpicture}
			\draw[gray] (0,0) -- (6,0);
		\end{tikzpicture}
		
	\end{subfigure}

	\vspace{.3em}
	
	\begin{subfigure}{.4\linewidth}
		\begin{flushright}
			\begin{tikzpicture}[scale=0.12]
				\draw[rounded corners=0.5mm,line width=0.5mm, fill=gray, fill opacity=0.1] (0,0)--(0,1)--(1,1)--(1,2)--(2,2)--(2,5)--(1,5)--(1,6)--(0,6)--(0,7)--(3,7)--(3,8)--(5,8)--(5,9)--(8,9)--(8,8)--(10,8)--(10,7)--(13,7)--(13,6)--(12,6)--(12,5)--(11,5)--(11,2)--(12,2)--(12,1)--(13,1)--(13,0)--(0,0);
			
			\draw[rounded corners=0.5mm,line width=0.5mm, fill=white] (3,2)--(4,2)--(4,1)--(5,1)--(5,2)--(6,2)--(6,5)--(5,5)--(5,6)--(4,6)--(4,5)--(3,5)--(3,2);
			
			\draw[rounded corners=0.5mm,line width=0.5mm, fill=white] (7,2)--(8,2)--(8,1)--(9,1)--(9,2)--(10,2)--(10,5)--(9,5)--(9,6)--(8,6)--(8,5)--(7,5)--(7,2);
			
			\draw[line width=0.3mm] (1,1)--(12,1);
			\draw[line width=0.3mm] (1,6)--(12,6);
			\draw[line width=0.3mm] (2,7)--(10,7);
			\draw[line width=0.3mm] (5,8)--(8,8);
			\end{tikzpicture}
		\end{flushright}
		
	\end{subfigure}\hspace{2em}
	\begin{subfigure}{.4\linewidth}
		\begin{flushleft}
			\begin{tikzpicture}[scale=0.13]
			\draw[rounded corners=0.5mm,line width=0.5mm, fill=gray, fill opacity=0.1] (0,0)--(11,0)--(11,1)--(9,1)--(9,4)--(10,4)--(10,6)--(8,6)--(8,7)--(6,7)--(6,8)--(5,8)--(5,7)--(3,7)--(3,6)--(1,6)--(1,4)--(2,4)--(2,1)--(0,1)--(0,0);
			
			\draw[rounded corners=0.5mm,line width=0.5mm, fill=white] (3,1)--(8,1)--(8,4)--(7,4)--(4,4)--(3,4)--(3,1);
			
			\draw[line width=0.3mm] (1,0)--(1,1);
			\draw[line width=0.3mm] (10,0)--(10,1);
			\draw[line width=0.3mm] (4,0)--(4,1);
			\draw[line width=0.3mm] (7,0)--(7,1);
			\draw[line width=0.3mm] (1,6)--(10,6);
			\draw[line width=0.3mm] (1,5)--(10,5);
			\draw[line width=0.3mm] (4,4)--(4,5);
			\draw[line width=0.3mm] (7,4)--(7,5);
			\draw[line width=0.3mm] (5,7)--(6,7);
			\draw[line width=0.3mm] (3,5)--(3,6);
			\draw[line width=0.3mm] (8,5)--(8,6);

		\end{tikzpicture}
		\end{flushleft}
		
	\end{subfigure}
	
	\caption{Complex Lego arcades (top, built with only two distinct Lego bricks: \brick{} and \hor{}) become easier to build after new, useful types of bricks are introduced (bottom). The complexity is measured as the number of pieces needed for the construction.}
	\label{fig:lego}

\end{figure}

%Restructuring makes knowledge more reusable, understandable, and efficient~\cite{Rumelhart1976AccretionTA}. % \ac{says who? Rumelhart and Norman?}.
The key to effective restructuring is finding appropriate abstractions.
As a running example, consider building Lego structures.
Figure \ref{fig:lego} (top) shows two structures built using only two types of bricks: short \brick{} and long \hor{}. % \ac{can we give them names, for instance, short and long?}
Building the structures using only these two types of bricks is complex and requires 29 and 18 bricks respectively.
However, as Figure \ref{fig:lego} (bottom) shows, by introducing new types of bricks through restructuring, such as a pillar and a horizontal brick of various lengths, we can build the same structures using only 7 and 11 pieces respectively.
In other words, by finding suitable abstractions, we can make the structures, and potentially future structures, easier and faster to build.

The importance of abstraction in AI is well-known \cite{saitta2013abstraction}.
However, the majority of learning AI agents merely accumulate knowledge, which is a problem because purely accumulating knowledge can be detrimental to learning performance \cite{srinivasan:2003,forgetgol}.
In other words, as knowledge is a form of inductive bias in machine learning \cite{mitchell:mlbook}, increasing the amount of knowledge increases the hypothesis space and consequently makes finding the target hypothesis more difficult.
The challenge is, therefore, to choose a learner's inductive bias (knowledge) so that the hypothesis space is large enough to contain the target hypothesis, yet small enough to be efficiently searched.

This paper aims to tackle the inductive bias problem by (i) reducing the size of the knowledge base, and (ii) restructuring it to make it easier to learn from.
Rather than only adding or removing knowledge \cite{DeRaedtLuc2008CpPp,mugg:metabias,forgetgol}, we argue that the human-like ability to \emph{restructure knowledge} can provide a better inductive bias to a learner and thus improve performance.
We call this problem \emph{knowledge refactoring}.
The idea is similar to program refactoring, where the goal of a programmer is to identify a good set of support functions to make a program more compact and reusable.

% A few approaches that tackle this challenge by revising knowledge do so to improve execution performance or make knowledge consistent with new observations~\cite{wrobel} \ac{this sentence breaks the flow. Can we cut because it is in related work?}

To restructure knowledge, we must explicitly store it.
This requirement eliminates non-symbolic learning approaches which dissipate knowledge in the parameters of a model.
We therefore use symbolic learning approaches, specifically inductive program synthesis \cite{shapiro:thesis}, which learns programs from input-output examples.
We focus on inductive logic programming (ILP) \cite{Muggleton94inductivelogic}, which represents background knowledge (BK) as a logic program and which has strong foundations in knowledge representation.

Our specific contributions are:
\begin{itemize}
    \item We introduce the \textit{knowledge refactoring} problem: revising a learner's knowledge base (a logic program) to reduce its size and minimise redundancy. Our key idea is to automatically identify useful substructures via predicate invention. The challenge lies in efficiently identifying substructures that lead to smaller programs. We tackle this challenge by casting the problem of knowledge refactoring as a constraint optimisation problem over a large set of candidate invented predicates.
    \item We introduce \knorf{}, a system that refactors knowledge bases by searching for new, reusable pieces of knowledge. %We cast the problem as a constraint optimisation problem (COP) ~\cite{CPHandbook}.
    A key feature of \knorf{} is that, rather than simply removing knowledge, it also introduces new knowledge through \emph{predicate invention} ~\cite{stahl}.
    \item We evaluate our approach on two domains: building Lego structures and real-word string transformations.
    Our experiments show that learning from refactored knowledge can substantially improve predictive accuracies of an ILP system and reduce learning times.
\end{itemize}

\section{Related Work}

% \textbf{Inductive program synthesis.}
% The goal of inductive program synthesis is to learn programs from input-output examples and background knowledge (BK).
% The goal of our work is to improve the performance of a program synthesis system by (1) reducing the size of the BK, and (2) restructuring the BK to make it easier to learn from.
%We focus on ILP, a form of program synthesis which learns programs from examples and BK, where the examples, BK, and programs are all logic programs.

\textbf{Redundancy elimination.}
Reducing redundancy is useful in many areas of AI, such as to improve SAT efficiency \cite{heule2015clause}.
In machine learning, irrelevant and redundant knowledge is detrimental to learning performance \cite{srinivasan:1995,srinivasan:2003,crop:reduce}.
Much work focuses on removing redundant literals or clauses from a logical theory \cite{plotkin:thesis}.
Theory \emph{minimisation} approaches try to find a minimum equivalent formula to a given input propositional formula \cite{DBLP:conf/ijcai/HemaspaandraS11} and also introduce new formulas.
By contrast, we focus on first on first-order (Horn) logic.
\emph{Forgetting} approaches \cite{forgetgol} try to remove clauses from the knowledge base to improve learning performance.
Our work is different because we (i) restructure knowledge, and (ii) introduce new knowledge through predicate invention.
% Whereas forgetting only removes clauses from the BK, we additionally introduce new knowledge to compress existing knowledge.

\textbf{Theory refinement.}
Theory \emph{refinement} \cite{wrobel} aims to improve the quality of a theory.
Theory \emph{revision} approaches \cite{DeRaedtRevision,AdeRevision,richards:mlj95} revise a program so that it entails missing answers or does not entail incorrect answers.
% (generalise it)
% (specialise it).
Theory \emph{compression} \cite{DeRaedtLuc2008CpPp} approaches select a subset of clauses such that the performance is minimally affected with respect to certain examples.
By contrast, our approach does not consider examples: we only consider the knowledge base.
Theory \textit{restructuring} changes the structure of a logic program to optimise its execution or its readability~\cite{wrobel}.
For instance, FENDER \cite{fender} restructures a theory with intra- and inter-construction operators~\cite{progol}.
The authors claim that their approach leads to a theory that is deeper, more modular, and possibly easier to understand and maintain.
By contrast, our goal is to restructure a theory by reducing the number of unnecessary predicate symbols in it and by introducing new ones.
 % which, as far as we are aware, is an original problem.
Moreover, we formulate the refactoring problem as a COP.

\textbf{Predicate invention.}
\knorf{} supports \emph{predicate invention} \cite{stahl}, the automatic introduction of new auxiliary predicates.
In contrast to the existing approaches which invent predicates before \cite{ijcai2019-841,DBLP:conf/ijcai/HocquetteM20} or during \cite{mugg:metagold} learning, \knorf{} invents them after learning through refactoring.
Three approaches are especially relevant to us.
Alps \cite{Alps} invents predicates by compressing a knowledge base formed of facts.
By contrast, \knorf{} considers and definite clauses with more than one literal.
% \ac{do you means facts rather than clauses with more than one literal?}
EC~\cite{ec} learns programs such that they are compressible, but does not revise previously invented abstractions, while \knorf{} does.
EC$^2$~\cite{Dreamcoder}, building upon EC, locally searches for small changes to a functional program to increase an optimisation function.
% (new $\lambda$-expressions)
Our approach differs because (i) we work in a purely logical setting, (ii) we preserve the semantics of the original program, and (iii) we solve the refactoring problem as a COP.

\section{Problem Description}
% Before introducing the knowledge refactoring problem we provide essential preliminaries on logic programming, after which we derive how knowledge refactoring can aide an ILP system such a Metagol.

To introduce the knowledge refactoring problem, we first provide essential preliminaries on logic programming (LP) \cite{reason:SteSha86a}, after which we show how knowledge refactoring can aide inductive program synthesis.
% To quantify the benefits of refactoring, we focus on the state-of-the-art ILP system Metagol \cite{metagol}.

\subsection*{Logic Programming}

\begin{figure}[t]
	\centering
		\begin{tikzpicture}[scale=0.25]
		
			% outline
			\draw[rounded corners=0.5mm,line width=0.5mm, fill=gray, fill opacity=0.1] (0,0) -- (3,0) -- (3,1) -- (2,1) -- (2,4) -- (3,4) -- (3,5) -- (0,5) -- (0,4) -- (1,4) -- (1,1) -- (0,1) -- (0,0);
			
			% pieces
			\draw[line width=0.3mm] (1,1)--(3,1);
			\draw[line width=0.3mm] (1,2)--(2,2);
			\draw[line width=0.3mm] (1,3)--(2,3);
			\draw[line width=0.3mm] (1,4)--(2,4);
			
			% ground
			\draw (-2,-0.2) -- (5,-0.2);
			\draw (-2,-0.2) -- (-2,-0.6);
			\draw (-1,-0.2) -- (-1,-0.6);
			\draw (0,-0.2) -- (0,-0.6);
			\draw (1,-0.2) -- (1,-0.6);
			\draw (2,-0.2) -- (2,-0.6);
			\draw (3,-0.2) -- (3,-0.6);
			\draw (4,-0.2) -- (4,-0.6);
			\draw (5,-0.2) -- (5,-0.6);
			
%			% cursor
%			\draw[-{Kite}] (0.5,-1) node (px) [left] {}; %{\tiny \code{X}};
%			\draw[-{Kite}] (1.5,-1.5) node (py) [right] {}; %{\tiny \code{Y}};
%			\draw[-{Kite}] (0.5,-2) node (pz) [left] {}; %{\tiny \code{Z}};
%			
			
			%\draw[-, thin] (0.5,-1) to (1.5,-1.5);

			\node at (-1.3,6) [align=left] {\footnotesize \code{Unfolded clause}};

			% outline
			
			\draw[rounded corners=0.5mm,line width=0.5mm, fill=gray, fill opacity=0.1] (10,0) -- (13,0) -- (13,1) -- (12,1) -- (12,4) -- (13,4) -- (13,5) -- (10,5) -- (10,4) -- (11,4) -- (11,1) -- (10,1) -- (10,0);

			%pieces
			\draw[line width=0.3mm] (11,1)--(12,1);
			\draw[line width=0.3mm] (11,4)--(12,4);
			
			%ground
			\draw (8,-0.2) -- (15,-0.2);
			\draw (8,-0.2) -- (8,-0.6);
			\draw (9,-0.2) -- (9,-0.6);
			\draw (10,-0.2) -- (10,-0.6);
			\draw (11,-0.2) -- (11,-0.6);
			\draw (12,-0.2) -- (12,-0.6);
			\draw (13,-0.2) -- (13,-0.6);
			\draw (14,-0.2) -- (14,-0.6);
			\draw (15,-0.2) -- (15,-0.6);
			
			\node at (13.5,6) [align=right] {\footnotesize \code{Folded clause}};

		\end{tikzpicture}
	% Tias: I replaced 'str' by 'shp1' as 'str' reminds of string where strings are the other task...
	\caption{Construction of the pillar (left) can be simplified by \textit{abstracting} (or \textit{folding} in LP) the procedure for constructing the vertical piece in the middle (right).}
	\label{fig:cl}

\end{figure}
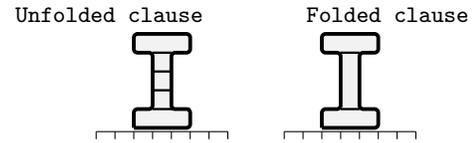

%\texttt{placeL(B) $\leftarrow$ place(B),right(B,C),place(C),}
%	
%	\texttt{onTop(C,D),place(D),onTop(D,E),place(D).}

A definite logic program is a set of definite clauses of the form \code{head \dep} \code{ cond$_1$, \ldots, cond$_N$.}
%\ac{we should be clear that we focus on definite programs, just in case the Imperial group read the paper!}
% \begin{align}
% \end{align}
A clause states that head is true if all conditions are true.
The head and conditions are \textbf{atoms} or their negations (jointly called literals), i.e., structured terms that represent relations between objects.
In the Lego example, \code{place(\brick{},Po,E,E$^\prime$)} is an atom, consisting of a predicate \code{place/4}, which \textit{places a brick of the type \brick{} at a position \code{Po} in a world with the state \code{E}, resulting in a new state \code{E$^\prime$}}.
Assume a one-dimensional world with Lego pieces and a cursor indicating the current position (Figure \ref{fig:cl}).
The clause:
\begin{lstlisting}[name=human,firstnumber=auto]
@\bluecode{pillar}(X,Y,E,E$^\prime$) \dep \\  place(\hor{},X,E,E$_1$), right(X,Z), place(\brick{},Z,E$_1$,E$_2$), place(\brick{},Z,E$_2$,E$_3$), place(\brick{},Z,E$_3$,E$_4$), left(Z,Y), place(\hor{},X,E$_4$,E$^\prime$).@ @\label{cl:unfolded}@
\end{lstlisting}
provides instructions for constructing a pillar at the position \code{X}: place a horizontal brick on position \code{X}, move the cursor to the position right of \code{X}, put three bricks on top of each other, move the cursor back to the starting position and place another horizontal brick (Figure \ref{fig:cl}, left).
%The final position of the cursor is \code{Y} and \code{E$^\prime$} is the state of the environment after all bricks were placed.

A key concept in LP is (un)folding \cite{unfolding}.
Intuitively, given a set of clauses \code{S}, the \code{fold(P,S)} operation replaces every occurrence of the body of clause \code{c} $\in \code{S}$, up to variable renaming, in program \code{P} with its head.
For instance, folding the clause \ref{cl:unfolded} with the clause:
\begin{lstlisting}[name=human,firstnumber=auto]
	@\bluecode{ver}(X,E,E$^\prime$) \dep \\ place(\brick{},X,E,E$_1$), place(\brick{},X,E$_1$,E$_2$), place(\brick{},X,E$_2$,E$^\prime$).@ @\label{cl:folding}@
\end{lstlisting}
results in the clause:
\begin{lstlisting}[name=human,firstnumber=auto]
	@\bluecode{pillar}(X,Y,E,E$^\prime$) \dep \\  place(\hor{},X,E,E$_1$), right(X,Z), \bluecode{ver}(Z,E$_1$,E$_2$), left(Z,Y), place(\hor{},X,E$_2$,E$^\prime$).@ @\label{cl:folded}@
\end{lstlisting}

\noindent
The \code{unfold(P)} operation essentially inlines all functions: for every clause \code{c} in program \code{P} which defines a predicate that is used in the body of another clause, it replaces every occurrence of the head of \code{c} in \code{P} with its body.
For instance, unfolding the clause \ref{cl:folded} with the clause \ref{cl:folding} results in the clause \ref{cl:unfolded}.
We assume that every inlined clause is removed from the program after unfolding.

\subsection{Knowledge Refactoring Problem}
\label{sec:refactoring}
% \ac{I would add a very high-level overview of the problem here saying what the input is and what the output is}

%Consider the folded program for pillar construction (clauses \ref{cl:folding} and \ref{cl:folded}; Figure \ref{fig:cl} right).
Consider our running example of constructing arcade structures (Figure \ref{fig:lego}).
The predicates in the logic program have different roles:

\begin{itemize}
    \item \textit{Primitive} predicates represent user-provided primitive knowledge that cannot be further decomposed, e.g. \code{place/4}, \code{right/2} and \code{left/2}.
	\item \textit{Task} predicates define solutions to tasks we want to solve or have solved, e.g. arcade structures.
    % Task predicates include predicates that only appear in the heads of clauses, as well as any recursive predicate.
    \item \textit{Support} predicates represent useful abstractions, e.g. \bluecode{ver}\code{/3}.
    They help us better structure a program but can be unfolded from a program without changing its semantics with respect to the task predicates.
    % , i.e. they disappear from the program once unfolded.
    We denote support clauses in \bluecode{blue} throughout the paper.
\end{itemize}

\noindent

Our refactoring problem takes as input a space of possible support predicates.
We restrict support clauses by their size.
The size of a clause \code{c}, \code{size(c)}, is the number of literals in \code{c} (including the head atom).
We define the support clause space:

\begin{definition}[\textbf{Support clause space}]
A clause $c$ is in the support clause space $\mathcal{S}$ of a program $P$ if (i) the head predicate of $c$ does not appear in $P$, and (ii) the predicates in the body of $c$ are in $P$ or other support clauses. % the predicate symbols in the body $c$ are primitives of $P$ or other support predicates. %,  and (iii) $size(c) \leq \max \bigcup_{d \in P} size(d)$.
%>>>>>>> 90c031ee5c994e4d1e763494802cf8951c448b56
\end{definition}

\noindent
When we refactor a program, we want to preserve the semantics of the original program with respect to complex predicates.
We reason about the restricted consequences of a program:

\begin{definition}[\textbf{Restricted consequences}]
Let $T$ be a set of predicate symbols and $P$ be a logic program.
The consequences of $P$ restricted to $T$ is $M_T(P) = \{ a | a \in atoms(P), P \models a, \text{the predicate symbol of a is in } T\}$.
\end{definition}

% \ac{give example}

\noindent
We also want to reduce the size of the original program.
The function $size(P)$ denotes the total number of literals in the logic program $P$.
We define the \emph{knowledge refactoring} problem:

\begin{definition}[\textbf{Knowledge refactoring}]
Let $P$ be a logic program, $T$ be a set of task predicate symbols, and $\mathcal{S}$ be a set of support clauses.
Then the refactoring problem is to find $P^\prime \subseteq \code{fold}(\code{unfold}(P), \mathcal{S})$ such that (i) $M_T(P^\prime)$ == $M_T(P)$, and (ii) $size(P^\prime) < size(P)$.
\label{def:refactoring}
\end{definition}

\noindent
This definition provides conditions for refactoring: it should yield  support clauses that, once folded into a program, (i) preserve the semantics of the  original program, and (ii) lead to the smaller program.
Refactoring therefore produces a lossless compression of the unfolded program, with respect to the main predicates.
Importantly, it leaves the construction of the support clauses open as there are many valid ways to do so.
We detail this aspect when discussing the implementation of the system.

\subsection{Benefit of Refactoring}
To show the potential benefits of refactoring, imagine an ILP system that enumerates all programs in the hypothesis space, a common approach when inducing functional programs ~\cite{deepcoder,Dreamcoder}.
Ignoring first-order variables for simplicity, the size of the hypothesis space is at most ${p^l \choose m}$ where $p$ is the number of predicate symbols allowed in a hypothesis, $l$ is the maximal number of literals in the body of a clause in a hypothesis, and $m$ is the maximum number of clauses in a hypothesis.
According to the Blumer bound \cite{behw-or-87}, given two hypothesis spaces of different sizes, and assuming that the target hypothesis is in both spaces, searching the smaller space will result in fewer errors compared to the larger one.
This result implies that we can improve the performance of an ILP system by either reducing the number of predicate symbols $p$ or the size of the target program $n$.
By refactoring we can reduce (i) $p$ by removing redundant predicate symbols and also by limiting the number of predicate symbols allowed in the BK, and (ii) $m$ and $l$ by restructuring the BK so that we can express the target hypothesis (program) using fewer, or shorter, clauses.
We argue that refactoring is especially important in a \textit{lifelong learning} setting where a system continuously learns thousands of new concepts, i.e. where $p$ can be very large.

\begin{figure*}[t]
    \centering
    \includegraphics[width=.9\linewidth]{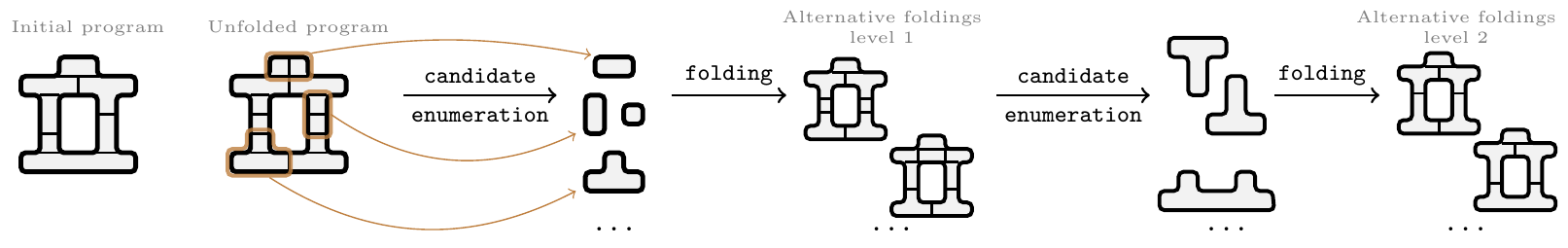}
    \caption{To identify candidate support clauses, \knorf{} first unfolds the original program. Then, \knorf{} constructs candidates from subparts of the unfolded program. To obtain more complex support clauses, \knorf{} first constructs alternative foldings of the given program, using the previously constructed candidates, and repeats the same procedure.}\label{fig:cands}
\end{figure*}

\section{\knorf{}: A Knowledge Refactoring System}

\subsection{Syntactic Refactoring}

The refactoring problem (Definition \ref{def:refactoring}) requires that the refactored program is (in a restricted form) semantically equivalent to the original program.
% That is, the refactored program should have the same success set, restricted to task predicates, as the original program.
However, checking this requirement is intractable in practice because we need to check that two programs produce the same output for every possible input, which could be infinite.
To make the problem tractable, \knorf{} uses a weaker criterion of \textit{syntactic equivalence}:%As syntactically equivalent programs are also semantically equivalent, \knorf{} solves the refactoring problem (Definition \ref{def:refactoring}).

%\ac{I do not understand this part.
%We state that ''semantic equivalence is intractable''.
%We then propose a method that is tractable yet is equivalent to 'semantic equivalence'.
%It seems contradictory.
%I will think how to improve it.
%}
% \seb{improved?}

\begin{definition}[\textbf{Syntactical equivalence}]
A program $P$ is syntactically equivalent to the program $P^\prime$ if $\code{unfold}(P) = \code{unfold}(P^\prime$).
\end{definition}

\noindent
Note that two syntactically equivalent programs are necessarily semantically equivalent, while the opposite does not hold.

As \knorf{} searches for the syntactically equivalent refactoring, it constructs the support clause space by extracting subsets of body literals from the unfolded original program.
We now introduce concepts necessary to construct the support clause space.

%\ac{we need to introduce these definitions. Four successive definitions without any helper text are very difficult to understand.}

\begin{definition}[\textbf{Connected clause}]
A clause $C$ is \textit{connected} if it cannot be partitioned into two non-empty clauses $C_1$ and $C_2$ such that the variables in $C_1$ are disjoint from $C_2$, i.e,
$\nexists C_1,C_2 \subseteq C \text{ such that } C_1 \neq C_2, \text{vars}(C_1) \cap \text{vars}(C_2) = \emptyset $.

%\begin{example}

%\end{example}

% $\nexists D \subseteq C \text{ such that } \text{vars}(D) \cap \text{vars}(C \setminus A) = \emptyset $.

% A clause $C$ is \textit{connected} if it cannot be partitioned into two clauses such that the variables appearing in one clause is disjoint from the other, i.e,
% $\nexists D \subseteq C \text{ such that } \text{vars}(D) \cap \text{vars}(C \setminus A) = \emptyset $.
% \ac{any reason for choosing the symbol $\mathcal{L}$? This symbol is often used to denote a language class, so it may be confusing.}
% \ac{a set of literals is a clause}

\end{definition}
\noindent
For instance, the clause \code{h(X,Y) \dep p(X,Y), q(Z)} is not a connected clause because the variables can be partitioned in two disjoined subsets, \code{\{X,Y\}} and \code{\{Z\}}.

\begin{definition}[\textbf{Clausal power-set}]
A \textit{clausal power-set} of the clause $c$, $\mathcal{P}(c)$, is the power-set of the literals in the body of $c$, excluding the empty set.
\end{definition}

%\begin{example}
\noindent
For instance, a clausal power-set of the clause \code{h(X,Y) \dep a(X,Y),b(Y,Z),c(Z).} is {\small \{\{\code{a(X,Y)}\}, \{\code{b(Y,Z)}\}, \{\code{c(Z)}\},\{\code{a(X,Y),b(Y,Z)}\}, \{\code{a(X,Y),c(Z)}\}, \{\code{b(Y,Z),c(Z)}\}, \{\code{a(X,Y),b(Y,Z),c(Z)}\}\}}. \label{ex:2}
%\end{example}

\begin{definition}{\textbf{(Connected clausal power-set)}}
A \textit{connected clausal power-set} of the clause $c$, $\mathcal{C}(c)$, is the maximal subset of $\mathcal{P}(c)$ such at every $s \in \mathcal{C}(c)$ is connected: $\mathcal{C}(c) = \{ e \in \mathcal{P}(c) | \text{ connected}(c)\}$.
In other words, only those subsets of literals of $c$ that are connected.
\end{definition}

%\begin{example}
\noindent
Continuing on the previous example, the connected clausal power-set removes \{\code{a(X,Y),c(Z)}\} from the clausal power-set  because the variables \{\code{X,Y}\} are disjoint from \{\code{Z}\}.
%\end{example}

With these concepts in place, we define the space of support clauses.

\begin{definition}[\textbf{Space of support clauses}]
A clause is in the support clause space $\mathcal{S}^i_j$ of a program $P$ when (i) it has at least $i$ and at most $j$ literals in the body, (ii) the set of literals in the body is in $\bigcup_{c \in P} \mathcal{C}(c)$ (up to variable renaming), and (iii) the head predicate symbol is unique and does not appear in $P$. \label{def:scs}
\end{definition}

\noindent
\knorf{} solves the refactoring problem by transforming it to a \textit{constraint optimisation problem} (COP) \cite{CPHandbook}, where the goal is to find an optimal set of support clauses.
Given (i) a set of \textit{decision variables}, (ii) a problem description in terms of \textit{constraints}, and (iii) an \textit{objective function}, a COP solver finds an assignment to decision variables that satisfies all specified constraints and maximises or minimises the objective function\footnote{We use the CP-SAT solver \cite{ortools}.}.

\knorf{} minimises both size and redundancy:
%\ac{How much better? I think many readers will want to know about this tricks. Perhaps we could say that, because of space limitations, we could not show these results in the paper, but will include them in any extended version (and give an indication for how much it reduces running time).} \seb{I completely agree, but do not have the results}

\begin{definition}[\textbf{Syntactic redundancy}]
A logic program $T$ has \emph{syntactic redundancy} if there are two clauses $c_1, c_2 \in T$ such that $c_1 \neq c_2$, $u_1 \in \mathcal{C}(c_1)$, $u_2 \in \mathcal{C}(c_2)$, $size(u_1) > 1$, $size(u_2) > 1$, and $u_1$ and $u_2$ are the same up to the variable renaming.
\end{definition}

\noindent
In other words, two clauses have a common subset of body literals.cp k	
Though minimising program size should imply the removal of redundancy, we notice empirically that minimising both better guides the search to good solutions, e.g. within a certain time limit

% \ac{To find the best set of support clauses, we build and order them constructively from simple to more complex. Does this fit here?}

\subsection{Decision Variables: Support Clauses}

\knorf{} solves the refactoring problem as a subset selection problem over the space of support clauses (Definition \ref{def:scs}).
%Therefore, the decision variables are Boolean variables which indicate whether a particular support clause is included in the refactored program.
In principle, the support clause space $\mathcal{S}^i_j$ is infinite, even with upper-bounded length of clauses, as any number of support predicates can be introduced.
To avoid this issue, we introduce an incremental procedure to construct $\mathcal{S}^i_j$ with clauses of fixed length.
The procedure repeatedly applies two steps, \textbf{candidate extraction} and \textbf{folding}, starting from the unfolded program \code{P}.
The unfolded program contains only \textit{primitives} (in Figure \ref{fig:cands} left, this results in clauses placing only \brick{} and \hor{} pieces).

The \textbf{candidate extraction} step  constructs support clauses from the connected power-sets of the clauses from the unfolded program \code{P}, $\bigcup_{c \in \text{\code{P}}} \mathcal{C}(c)$, with at least $i$ and at most $j$ literals.
More specifically, \knorf{} turns each element of $\bigcup_{c \in \text{\code{P}}} \mathcal{C}(c)$ into a support clause by creating a new predicate symbol in the head.
These support clauses are expressed in terms of primitive predicates.
In the Lego example, taking $i=1$ and $j=2$ would result in some of the  candidates illustrated in Figure \ref{fig:cands}, middle.

The \textbf{folding} step folds the extracted support clauses into the (unfolded) program.%, until the bodies of all clauses are expressed entirely in terms of extracted candidates (each clause could have multiple foldings).
This step essentially rewrites the program such that the bodies of its clauses are made of support predicates (a single clause can have multiple foldings).
In the Lego example, this results in the simplified construction of the pillar structure (Figure \ref{fig:cands}, middle).
To obtain more complex support clauses, \knorf{} repeats the same two steps but starts from the folded program.

\knorf{} repeats these two steps until each clause in the program has only one body literal.
The result of this procedure is a \textbf{hierarchy} of support clauses, each one building on simpler support clauses.
We refer to these steps as \textit{levels of refactoring}; the folding the unfolded program yields \textit{level one} refactoring, folding again yields \textit{level two} refactoring, and so on.

Each enumerated support clause candidate \code{k} is associated with a Boolean variable \bluecode{sc}$_\code{k}$ indicating whether the support clause is selected.
Each folding of the clause \code{i} is associated with a Boolean variable $\code{f}^\code{i}_n$ indicating that a particular folding is selected as a part of the refactored program.

\subsubsection{Pruning Support Clauses}

The incremental candidate enumeration procedure described above still results in many candidates because each clause can be expressed in many ways, given a set of support clauses.
We further prune the support clause space by (i) eliminating singleton clauses, and (ii) removing clauses that cannot reduce the program size.

We remove support clauses with \textit{singleton variables}, i.e. clauses with a variable that only appear once.
For instance, the clause:
\begin{lstlisting}[name=human,firstnumber=auto]
	@\bluecode{sup}(X,E) \dep \\ place(\brick{},X,E,E$_1$), place(\brick{},X,E$_1$,E$^\prime$).@ @\label{cl:pruning1}@
\end{lstlisting}
is removed because \code{E$^\prime$} appears only once.
Adding \code{E$^\prime$} as the last argument in the head would make the clause valid.
%is a valid candidate and is not removed because all variables appear at least twice
%If the variable \code{E$^\prime$} would not appear in the head of the clause, such clause would be removed as \code{E$^\prime$} appears only once.
%, while the clause:
%\begin{lstlisting}[name=human,firstnumber=auto]
%	@\bluecode{sup2}(X,E) \dep \\ place(\brick{},X,E,E$_1$), place(\brick{},X,E$_1$,E$_2$), place(\brick{},X,E$_2$,E$^\prime$).@ @\label{cl:pruning2}@
%\end{lstlisting}
%is removed because \code{E$^\prime$} only appears once.
As we focus on inductive program synthesis problems, ignoring singleton clauses is not sacrificing expressivity because singleton variables are essentially variables that are never used.

We also remove support clauses that cannot reduce the size of the program.
For instance, let \code{c} be a support clause and \code{usage(c, \cal{T})} be the number of clauses in the program \code{T} which can be folded with \code{c}. %where $c$ could be used, i.e. $usage(c, \mathcal{T}) = |\{t \in \mathcal{T} | c \subseteq \mathcal{C}(t)\}|$.
This means that in the best case, we can replace \code{usage(c, T)}$\times$(\code{size(c)}$-1$) literals in the program (i.e., every occurrence of the body of \code{c}) by \code{usage(c, T)}$\times1$ uses of the head of the support clause \code{c} and the addition of a clause \code{c} to the theory \code{T}.
Hence, if it is the case that this inequality holds: {\small $$\text{\code{usage(c, T)}$\times$(\code{size(c)}$-1$) $\leq$ \code{usage(c, T)} + \code{size(c)}}$$}
\noindent
Then we know that the use of this candidate support clause will never lead to a reduction in the program size (our overall goal). %hence it should not be considered as a candidate support clause.
We remove candidate support clauses that violate this inequality.

\subsection{Constraints: Valid Refactoring}
\label{cons:vr}

Each clause in the unfolded program has multiple possible foldings, grouped in different levels due to the support clause generation process.
The refactored program should replace the original clauses with one of the possible foldings.
Hence, \knorf{} enforces a constraint stating that at least one of the foldings of the clause \code{i} should be formed by the chosen support clauses.

We group the foldings of the clause \code{i} per level and add an additional level indicator $\code{l}^\code{i}_d$.
For reasons that will become obvious later, the level indicator $\code{l}^i_d$ is a Boolean variable which is set to \code{true} if the selected folding of the clause \code{i} comes from the level $d$.
%Assume that $\code{f}^\code{i}_n$ is a Boolean variable indicating that the $n$-th foldings of clause \code{i} can be constructed with the given set of candidate support clauses.
This results in constraints of the form:
{\small $$\bigvee_{d=1}^{\text{max levels}} \ \left( \code{l}^i_d \wedge \left( \bigvee_{n} \code{f}^i_n \right) \right).$$}
%where $\texttt{l}^i_l$ is \textit{true} if level $l$ is the selected folding of clause $i$, and $\texttt{f}^i_n$ is a Boolean variable indicating that the $n$-th foldings of clause $i$ can be constructed with the given set of candidate support clauses.
%where $\code{f}^\code{i}_n$ is a Boolean variable indicating that the $n$-th foldings of clause \code{i} can be constructed with the given set of candidate support clauses.

\noindent
\knorf{} forces that one level of refactoring is chosen for each clause by imposing the following constraint:
{\small $$\sum_{d=1}^{\text{max levels}} \code{l}^i_d = 1. $$}
This level variable will be part of the objective, where higher levels are typically better.

To decide whether a folding $\code{f}^\code{i}_n$ can be constructed, the solver needs to know which support clauses are needed for that particular folding.
For instance, to construct the top folding at the level 1 in Figure \ref{fig:cands}, we need the following pieces: \brick{}, \horTwo{}, \pillarBottom{}, and \pillarTop{} (assume that the selection of these support clauses is indicate with the variables \bluecode{sc}$_p$, \bluecode{sc}$_r$ and \bluecode{sc}$_q$).
To ensure this connection, \knorf{} enforces the constraint stating that the folding $\code{f}^\code{i}_n$ can be constructed only if all the necessary pieces are selected as a part of the solution:
{\small $$ \code{f}^\code{i}_n \Leftrightarrow \left( \bluecode{sc}_p \wedge \bluecode{sc}_r \wedge \bluecode{sc}_q \right).$$}

\noindent
Finally, candidates extracted from  level $L$ depend on the candidates from the level $L-1$ (i.e., the bodies of support clauses from level $L$ are composed from the predicates introduced by the support clauses at level $L-1$).
\knorf{} imposes the constraint directly materialising this dependency -- if the support clause \bluecode{sc}$_k$ is selected as a part of the solution, then all support clauses defining the predicates in the body of the clause \bluecode{sc}$_k$ (assume \bluecode{sc}$_l$ and \bluecode{sc}$_{m}$ ) also have to be a part of the solution
{\small $$\bluecode{sc}_k \Rightarrow ( \bluecode{sc}_l \wedge \bluecode{sc}_{m}).$$}
For instance, one needs \hor{} and \brick{} bricks to make \pillarBottom{}.

\subsection{Objective: Size and Redundancy}
%\ac{I found this section difficult to read but I am unsure how to improve it}

\knorf{} searches for the smallest refactored program, syntactically equivalent to the original program, with the least redundancy.
The size of the refactored program equals the number of literals in it, i.e., the size of the \textit{selected} foldings and support clauses.
To guide a COP solver towards small refactored program, \knorf{} minimises the following objective function
{\small
$$ \underbrace{\sum_{\code{cl} \in \code{T}}  \sum_{d=1}^{\text{L}} \sum_{n = 1}^{F^d(\code{cl})} \code{size}(\code{f}_n^\code{cl})*\code{f}_n^\code{cl}*\code{l}_d^\code{cl}}_{\text{size of selected foldings}} + \underbrace{\sum_{\bluecode{sc} \in \mathcal{S}^i_j} \code{size}(\bluecode{sc})*\bluecode{sc}}_{\text{size of selected support clauses}}$$
}
\noindent where $L$ is the maximal number of level and $F^d(\code{cl})$ is the number of foldings of the clause \code{cl} at level $d$.
%\tias{this is not equivalent to program size: all valid foldings of a level are selected... also, it is a quadratic objective which makes it harder to solve. We will work on a linear set-covering formulation together} % please leave this comment for future reference

The level indicators introduced in Section \ref{cons:vr} are important to measure the program size correctly.
%Our formulation of the refactoring program implicitly assumes that a folding is constructed if all necessary support clauses are available.
Assume that the selected folding of a certain (unfolded) clause is at the level 3.
As any folding at level 3 is constructed from support predicates introduced at level 2, at least one folding at level 2 is possible to construct.
But any folding from level 2, if the selected one is at level 3, should not contribute to the size of the refactored program.
The level indicators ensure that only the selected folding contributes to the program size by multiplying the size of folding from lower levels by 0.

To minimise the redundancy between clauses, \knorf{} keeps track of all foldings that share literals in the body.
%For instance, clauses \ref{cl:pruning1} and \ref{cl:pruning2} have a shared subpart (up to variable renaming):
%\begin{lstlisting}[name=human,firstnumber=auto,numbers=none]
%	@place(\brick{},X,E,E$_1$), place(\brick{},X,E$_1$,E$^\prime$).@
%\end{lstlisting}
We then introduce a new Boolean variable (e.g., \code{r}$_i$) indicating whether more than one folding (e.g., corresponding to variables \code{f}$^\code{i}_n$ and \code{f}$^k_m$) with such redundancy can be constructed
%
%To include the redundancy count in the objective function, we would introduce a new boolean variable \texttt{r}$_1$ which takes the \textit{true} value only if both foldings can be constructed
{\small $$ \code{r}_i \Leftrightarrow \left ( \code{f}^\code{i}_n + \code{f}^\code{k}_m  > 1\right ).$$}
%$$ \texttt{r}_1 \Leftrightarrow \textfig{figs/stndalone_lego_init}{5ex} \ \wedge \ \textfig{figs/standalone_lego_init2}{4.5ex} $$
\knorf{} introduces such constraint for all found redundancies and adds the sum over \code{r} variables to the objective function.
%\tias{the number of possible pairs is gigantic? especially at low levels?}

\section{Experiments}

%\seb{To add:
%\begin{itemize}
%	\item Baseline that identifies all redundancies in a theory an replaces them with a new predicate
%	\item correct timings
%	\item learning curves
%	\item graphs for up to 4000 background relations
%\end{itemize}
%}

We argue that an ILP system can learn better from refactored BK.
Our experiments therefore aim to answer the question:
\begin{itemize}
    \item[\textbf{Q:}] Can an ILP system learn \emph{better} with refactored BK?
\end{itemize}

\noindent
By better, we ask whether it can solve more tasks, learn with higher predictive accuracies, or learn in less time.
To answer this question, we compare the performance of state-of-the-art ILP system Metagol \cite{metagol} with and without refactored BK.

\textbf{Lifelong learning.}
To evaluate the usefulness of refactoring, we focus on a lifelong learning scenarios in which a learner continuously learns to perform new tasks by continuously adding programs to its BK.
This allows us to evaluate the benefit of refactoring over BKs with various sizes.
To generate the BK in this setting, we use Playgol~\cite{ijcai2019-841}\footnote{
The original work performs simple deduplication of clauses.
To fully verify the usefulness of refactoring, we have disabled this step.
}, an ILP system that generates BK automatically.
Playgol learns in two phases.
In the first \textbf{play} phase, Playgol solves randomly generated tasks that are similar to the user-provided target tasks.
In the second \textbf{build} phase, Playgol solves the user-provided tasks, using the solutions to the play tasks as BK.
We refer to the play tasks as \textit{background tasks} and generate BKs with $n$ background tasks, $n \in \{200, 400, \ldots, 4000\}$
%We evaluate Metagol when learning to solve user-provided tasks from (i) the generated BK (\textbf{No Refactoring}), and (ii) the BK after refactoring, i.e. after \knorf{} has refactored it (\textbf{Refactoring}).

\textbf{Systems.}
We evaluate Metagol when learning to solve user-provided tasks from (i) the generated BK (\textbf{No refactoring}), (ii)  the BK after refactoring, i.e. after \knorf{} has refactored it (\textbf{Refactoring}), and (iii) the BK refactored with a simple form of refactoring that replaces very redundancy in a program with a new predicate symbol and represents the redundancy with an additional clause (\textbf{No redundancy}).

\textbf{Experiment setting.}
To build the support clause space, we set the minimum and maximum length of support clauses to 2 and 3 respectively.
We impose no limit on the number of layers.
When solving the COP, we impose a timeout of 90 minutes.
If refactoring takes longer, we stop the search and take the best solution found so far.
We additionally impose a constraint that the refactored BK cannot have more predicates than the original BK.
We give Metagol a learning timeout of 60 seconds per task.
We repeat each experiment 10 times, and plot the means and 95\% confidence intervals.
All experiments are run on a CPU with 3.20 GHz and 16 Gb RAM.
We have allowed CP-SAT so use 8 parallel threads.

\begin{figure*}[t]
	\begin{minipage}[t]{0.32\linewidth}
		\begin{subfigure}{0.49\linewidth}
			\centering
			\includegraphics[width=\linewidth]{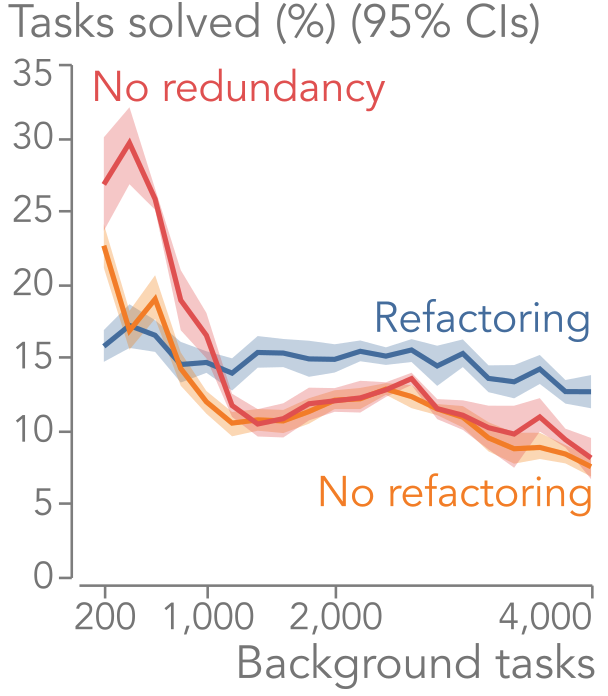}
			\caption{} \label{fig:lego:tasks}
		\end{subfigure}
		\begin{subfigure}{0.49\linewidth}
			\centering
			\includegraphics[width=\linewidth]{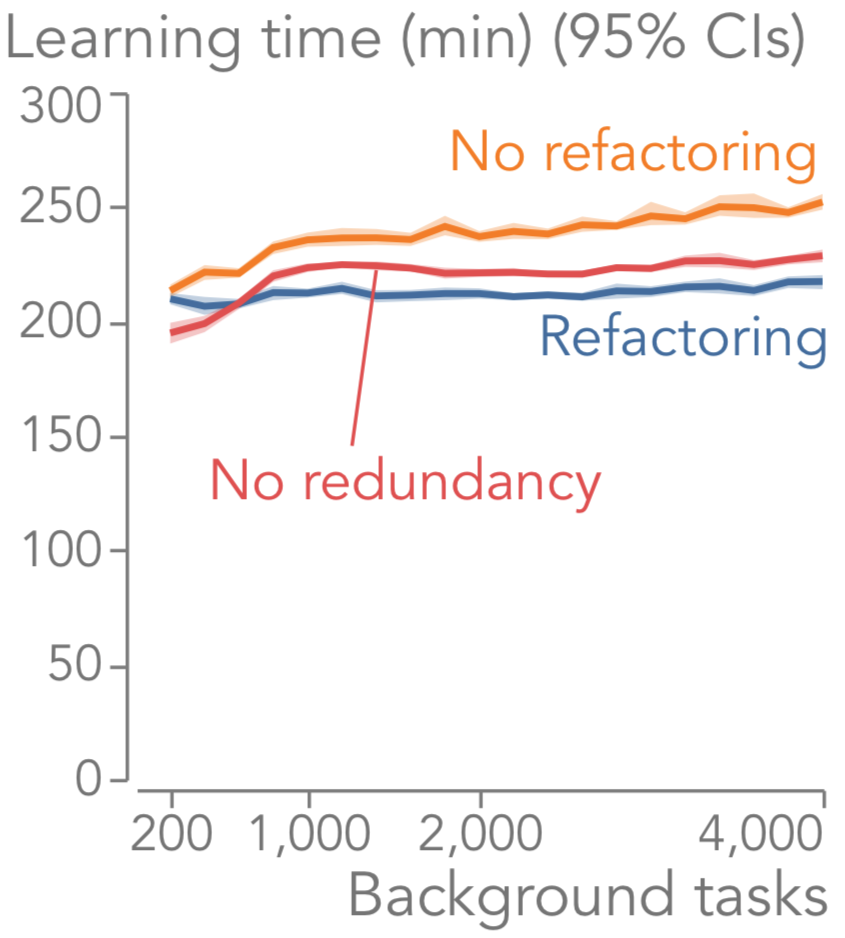}
        	\caption{}\label{fig:lego:time}
		\end{subfigure}
		\caption{With refactoring, Metagol solves more Lego tasks and does so in less than time than without refactoring.} \label{fig:lego:res}
	\end{minipage}
	\hfill
	\begin{minipage}[t]{0.32\linewidth}
		\begin{subfigure}{.49\linewidth}
			\centering
			\includegraphics[width=\linewidth]{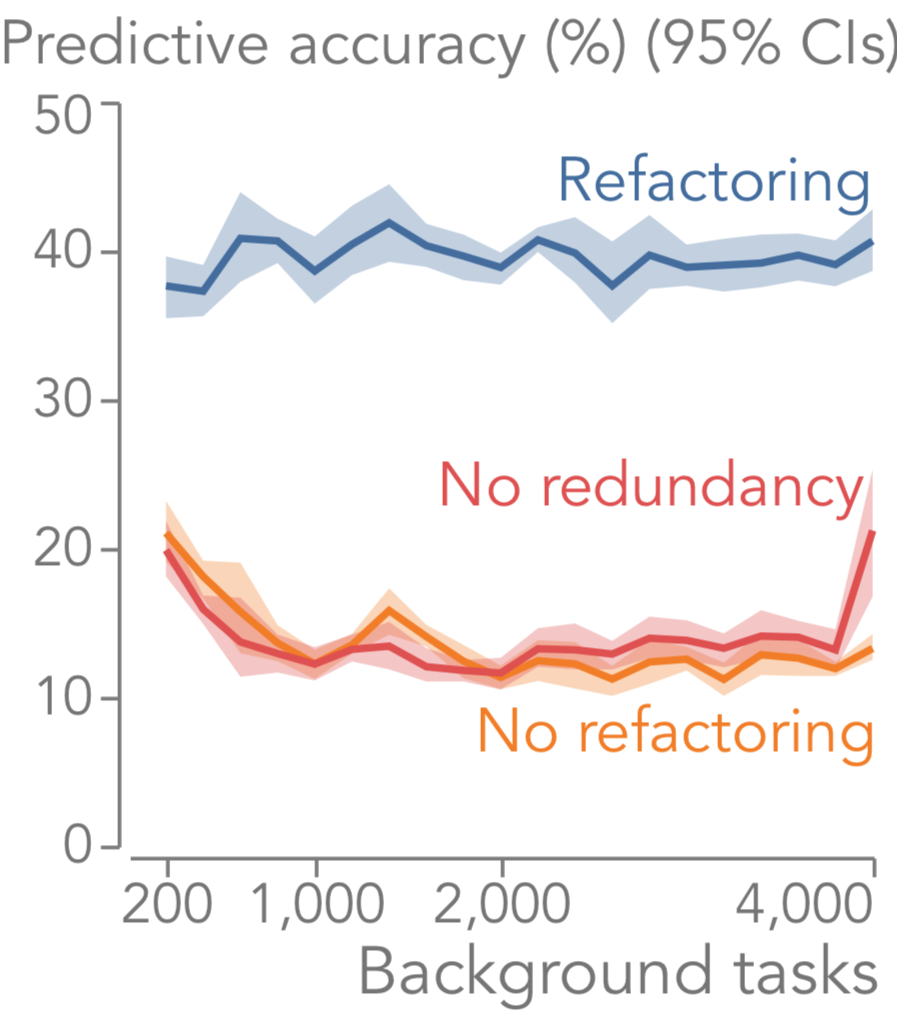}
        	\caption{} \label{fig:strings:tasks}
		\end{subfigure}
		\begin{subfigure}{.49\linewidth}
			\centering
			\includegraphics[width=\linewidth]{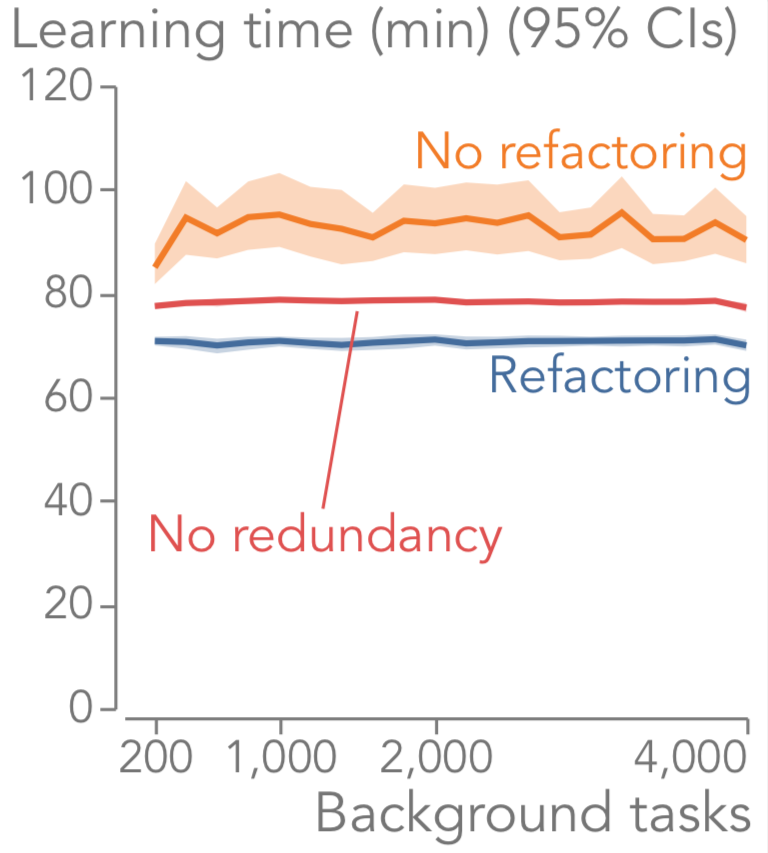}
        	\caption{}\label{fig:strings:time}
		\end{subfigure}
		\caption{With refactoring, Metagol solves more string transformation tasks and does so in less than time than without refactoring.} \label{fig:strings:time}
	\end{minipage}
	\hfill
	\begin{minipage}[t]{0.32\linewidth}
		\begin{subfigure}{.49\linewidth}
			\centering
			\includegraphics[width=\linewidth]{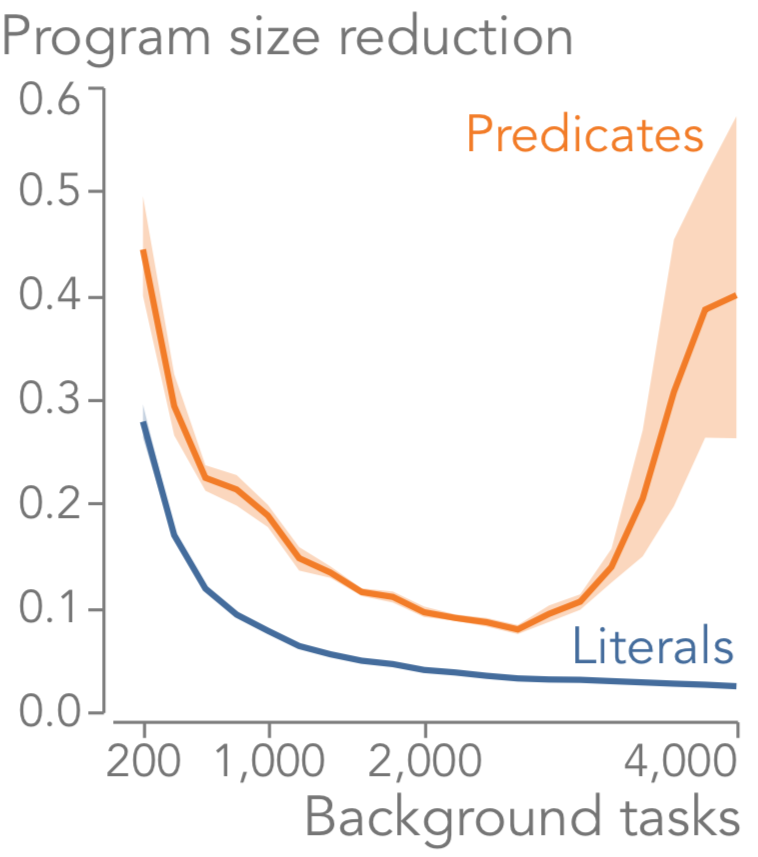}
        	\caption{Lego} \label{fig:lego:size}
		\end{subfigure}
		\begin{subfigure}{.49\linewidth}
			\centering
        	\includegraphics[width=\linewidth]{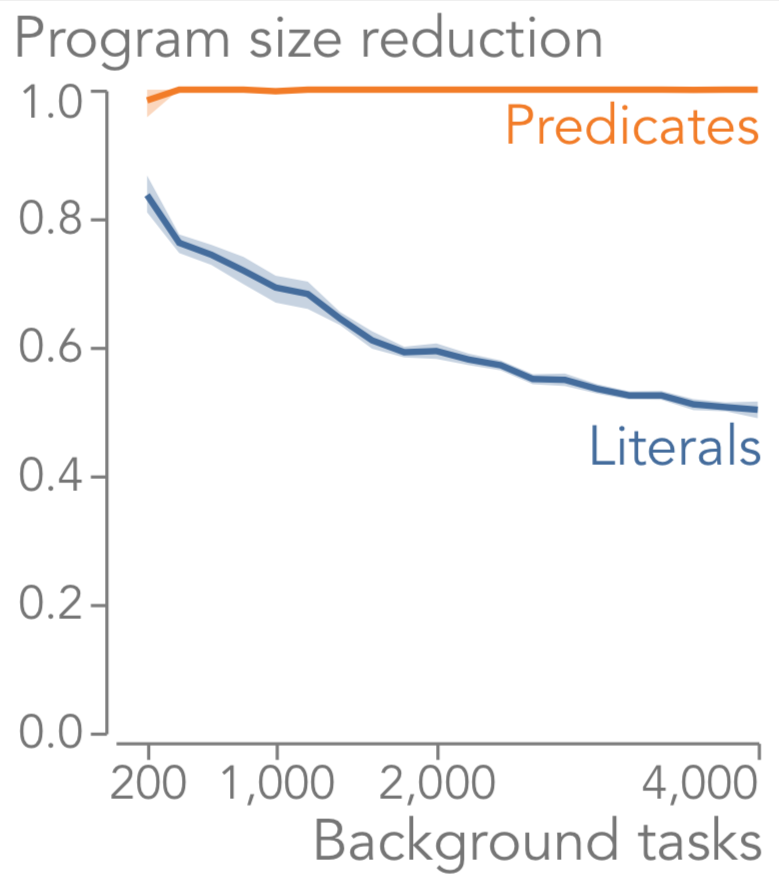}
        	\caption{Strings} \label{fig:strings:size}
		\end{subfigure}
		\caption{Refactoring reduces (shown as $\sfrac{\text{refactored}}{\text{original}}$) the number of predicates and literals in the program} \label{fig:size}
	\end{minipage}
\end{figure*}

\subsection{Experiment 1 - Lego}

Our first experiment is on learning to build Lego structures in a controlled environment \cite{forgetgol}. %In the next section we look at real-world string transformation learning.

%\begin{figure}[t]
%    \centering
%    \begin{subfigure}{.49\linewidth}
%        \centering
%        \includegraphics[width=.9\linewidth]{results/aaai21/lego/performance}
%        \caption{Percentage of tasks solved} \label{fig:lego:tasks}
%    \end{subfigure}
%    \begin{subfigure}{.49\linewidth}
%        \centering
%        \includegraphics[width=.9\linewidth]{results/aaai21/lego/runtime_trial}
%        \caption{Learning time}\label{fig:lego:time}
%    \end{subfigure}
%
%    \caption{With knowledge refactored, learning performance on Lego domain degrades more gracefully with the increase of background knowledge. Refactoring also reduces learning times.} \label{fig:lego:res}
%\end{figure}

\subsubsection{Materials}
We consider a Lego world with a base dimension of $6 \times 1$ on which bricks can be stacked.
We only consider $1 \times 1$ bricks of a single colour.
A training example is an atom $f(s1, s2)$, where $f$ is the target predicate and $s1$ and $s2$ are initial and final states respectively.
A state describes a Lego structure as a list of integers.
The value $k$ at index $i$ denotes that there are $k$ bricks stacked at position $i$.
The goal is to learn a program to build the Lego structure from a blank Lego board (a list of zeros).
We generate training examples by generating random final states.
The learner can move along the board using the actions \emph{left} and \emph{right}; can place a Lego brick using the action \emph{place\_brick}; and can use the fluents \emph{at\_left} and \emph{at\_right} and their negations to determine whether it is at the leftmost or rightmost board position.

\subsubsection{Method}
The background tasks were generated with a Lego board of size 2 to 4.
We randomly generate 1000 target tasks for a Lego board of size 6.
%We compare Metagol's learning performance on the target tasks with and without refactoring of the BK.
% We enforce a timeout of 60 seconds per task.
We measure the percentage of tasks solved (tasks where the Metagol learns a program) and learning times (total time need to solve all target tasks).

\subsubsection{Results}
%\ac{We cannot read the labels on the y-axes of the plots.}
The results (Figure \ref{fig:lego:tasks}) show that refactoring helps Metagol to maintain the performance when confronted with large BK.
With refactored BK, Metagol's performance decreases less with the increase of background tasks.
The results also show that refactoring slightly degrades the ability to solve tasks when BK is small.
The likely explanation is that smaller BK has less chance for redundancy and, thus, refactoring is eliminating predicates that Metagol could use to solve tasks.
When the BK is large ($\geq 1000$ background tasks), refactoring improves the ability to solve tasks.
These results appear to corroborate existing results \cite{forgetgol} which show that simple forgetting can improve learning performance but only when learning from lots of BK.
Figure \ref{fig:lego:time} also shows that refactoring reduces learning times by approximately 20\%.
Interestingly, refactoring by replacing redundancies (No redundancy) consistently reduces total learning times, but does not improve performance for a large BK.
% Similarly to the first experiment, \knorf{'s} performance is rather stable in contrast to learning from raw BK.

Figure \ref{fig:lego:size} shows that refactoring drastically reduces the size of the BK.
Both the number of literals and the number of predicates in the refactored BK is only a fraction of their number in the original BK.
This suggests that much of the raw BK obtained by Playgol can be represented using a shared set of a-priori unknown support clauses.

\subsection{Experiment 2 - String Transformations}

Our second experiment is on \emph{real-world} string transformations.
% , a domain commonly used to evaluate program induction systems \cite{mugg:metabias,deepcoder,ellis:repl}.

\subsubsection{Materials}
We use 130 string transformation tasks from \cite{ijcai2019-841}.
Each task has 10 examples.
An example is an atom $f(x,y)$ where $f$ is the task name and $x$ and $y$ are input and output strings respectively.
The goal is to learn to map the inputs to the outputs, such as to map the full name of a person (input) to its initials (output), e.g. \emph{'Alan Turing'} $\mapsto$ \emph{'A.T.'}.
We provide as BK the binary predicates \emph{mk\_uppercase}, \emph{mk\_lowercase}, \emph{skip}, \emph{copy}, \emph{write}, and the unary predicates \emph{is\_letter}, \emph{is\_uppercase}, \emph{is\_space}, \emph{is\_number}.

\subsubsection{Method}
We follow the procedure described in \cite{ijcai2019-841} to obtain the play tasks and thus BK.
For each of the 130 tasks, we sample uniformly without replacement 5 examples as training examples and use the remaining 5 as test examples.
We measure predictive accuracy and learning times (total time needed to solve all target tasks).

%\begin{figure}[t]
%    \centering
%    \begin{subfigure}{.49\linewidth}
%        \centering
%        \includegraphics[width=.9\linewidth]{results/aaai21/strings/performance}
%        \caption{Percentage of tasks solved} \label{fig:strings:tasks}
%    \end{subfigure}
%    \begin{subfigure}{.49\linewidth}
%        \centering
%        \includegraphics[width=.9\linewidth]{results/aaai21/strings/runtime_trial}
%        \caption{Learning time}\label{fig:strings:time}
%    \end{subfigure}
%
%    \caption{With knowledge refactored, Metagol solves more string transformation tasks. It also achieves so in less time.}
%    \label{fig:strings:res}
%\end{figure}

\subsubsection{Results}
The results (Figure \ref{fig:strings:tasks}) show that refactoring drastically improves predictive accuracies.
When learning from unrefactored BK and BK with redundancies removed, Metagol's performance quickly deteriorates because Metagol's search space increases exponentially in the size of the BK.
% That is furthermore emphasised by redundancy in BK, as Metagol cannot see that certain clauses might represent the same knowledge.
By contrast, when given refactored BK, Metagol has higher predictive accuracy in all cases, eventually four times higher than without refactored BK.
Moreover, the results shows that refactoring reduces learning times by a third.
Interestingly, the gain in performance does not come from the reduced number of predicates, as both refactored and unrefactored programs have equal number of predicates, though overall program size decreases (Figure \ref{fig:strings:size}).
Rather, the performance gains (both in accuracy and speed) come from \textbf{better structured} knowledge.

%\ac{We need to emphasise this point - it is finding better ways to represent knowledge}.

%\begin{figure}[t]
%    \centering
%    \begin{subfigure}{.49\linewidth}
%        \centering
%        \includegraphics[width=.9\linewidth]{results/aaai21/lego/program_reduction}
%        \caption{Lego problems} \label{fig:lego:size}
%    \end{subfigure}
%    \begin{subfigure}{.49\linewidth}
%        \centering
%        \includegraphics[width=.9\linewidth]{results/aaai21/strings/program_reduction}
%        \caption{String transformation} \label{fig:strings:size}
%    \end{subfigure}
%    \caption{Reduction in program size}
%    \label{fig:size}
%
%\end{figure}

\subsubsection{Solver Behaviour}
\begin{figure}
	\centering
	\includegraphics[width=.9\linewidth]{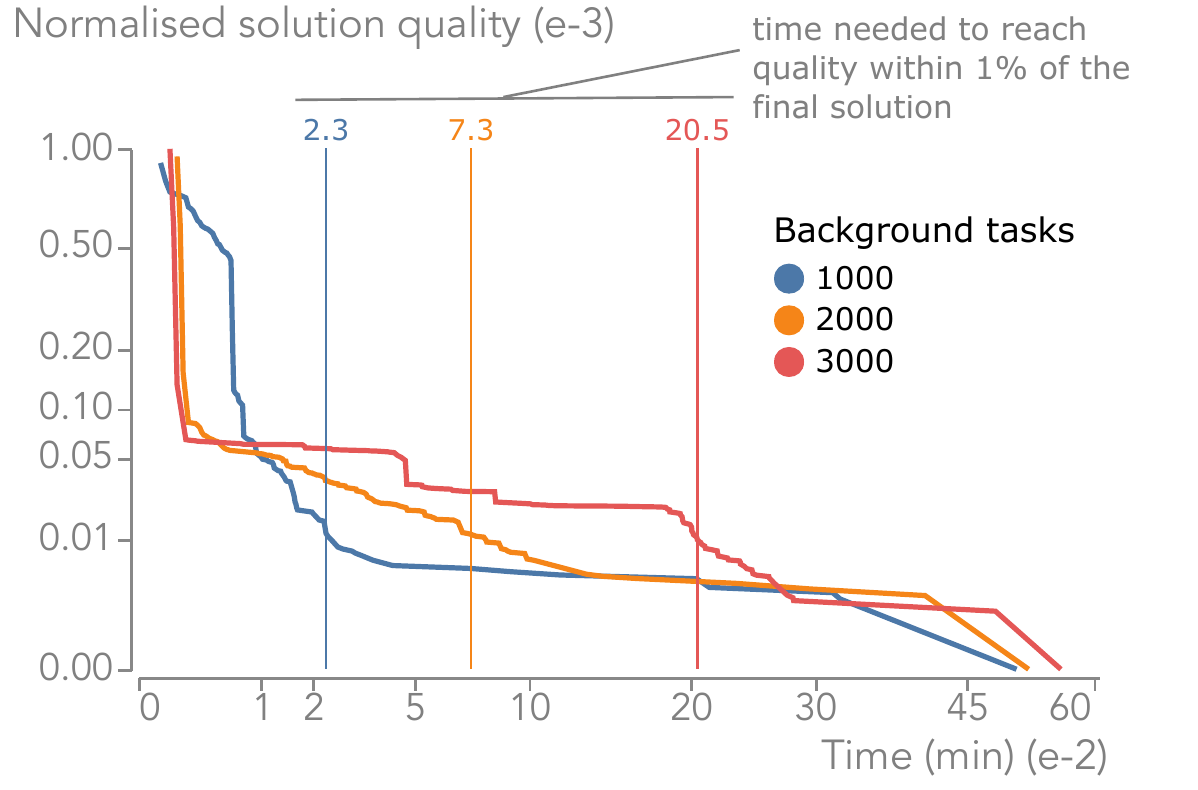}
	\caption{
    Despite the timeout of 90 min, the solver can quickly find a good solution and finds the best solution within an hour.
     Note that axes follow power scale.}
	\label{fig:solqual}
\end{figure}

Figure \ref{fig:solqual} shows the reduction of the BK size during a typical refactoring process for three different BK sizes.
Regardless of the size, the solver is able to find a good solution (within 10\% of final size) within a minute.
It takes between 2-21 min to reach a solution within 1\% of the final size, depending on the number of background tasks.
Though the solver find the best solution within an hour, for most of the runs it continues searching for a better solution until timeout.
This indicates two things: (1) we could have obtained equally good solutions with a more restrictive timeout, and (2) the encoding of a problem could be improved as the solver currently spends most of the time finding small improvements.

% Despite a large timeout, the solver is able to quickly find good solutions (

\section{Conclusion}

The main claim of this work is that the structure of an agent's knowledge can significantly influence its learning abilities: more knowledge results in larger hypothesis spaces and makes learning more difficult.
Focusing on inductive logic programming, we introduced a problem of knowledge refactoring -- rewriting an agent's knowledge base, expressed as a logic program, by removing the redundancies and minimising its size.
We also introduced \knorf{}, a system that performs automatic knowledge refactoring by formulating it as a constraint optimisation procedure.
We evaluated the proposed approach on two inductive program synthesis domains: building Lego structures and real-world string transformations.
Our experimental results show that learning from the refactored knowledge base results can increase predictive accuracies in fourfold and reduced learning times substantially.

%Humans are not mere fact accumulators; they revise and improve their knowledge to use it more efficiently.
%However, AI agents continuously learning from their environment do act as plain knowledge accumulators: they learn how to solve various tasks but never look back and introspect whether a newly discovered knowledge can help solve previously learnt tasks better.
%This work investigates how to give artificially intelligent agents the ability of revision.
%Focusing on program induction and Inductive Logic Programming, we introduce a problem of \textit{knowledge refactoring} -- rewriting the agents knowledge base, expressed as a logic program, so that the program size is minimal with little redundancy -- and investigate whether it can help agents learn more efficiently (either by learning faster or learning to solve more tasks).
%We also introduce \knorf{}, a system that performs automatic knowledge refactoring by formulating it as a constraint optimisation procedure.
%We evaluate the proposed approach on two program induction domains, string transformations and Lego structures, and show that learning from the refactored background knowledge resulting in more solved tasks in less time.

\subsubsection*{Limitations and Future Work}
We have used one ILP system that already performs predicate invention.
It would be interesting to see how effective refactoring is when a system that does not perform predicate invention.
We have focused on eliminating redundancy to improve performance of an ILP system.
However, there are many other properties that we may want to optimise, such as modularity or readability.
Finally, we have not tackled the question of \textit{when to refactor?}
Refactoring too often is likely to have negative effect on learning times.
We will investigate the strategies for detecting the need for refactoring in future work.
%Finally, we have tackled refactoring as a global problem.
%However, it might be possible to decompose the refactoring problem into simpler subproblems, which would allow us to (i) tackle larger programs, and (ii) explore more complex support clause spaces.

\begin{quote}
\begin{small}
\bibliographystyle{aaai}
\bibliography{knorf.bib}
\end{small}
\end{quote}

\end{document}